%% file: main.tex
\documentclass[10pt,twocolumn,letterpaper]{article}

\usepackage{iccv}    
\usepackage{array}
\usepackage{multicol}
\usepackage{multirow}
\usepackage{algorithm}
\usepackage{algpseudocode}
\usepackage{colortbl}
\usepackage{color, soul}


\definecolor{iccvblue}{rgb}{0.21,0.49,0.74}
\usepackage[pagebackref,breaklinks,colorlinks,allcolors=iccvblue]{hyperref}

\def\paperID{6} 
\def\confName{ICCV}
\def\confYear{2025}

\title{Med-GRIM: Enhanced Zero-Shot Medical VQA using prompt-embedded Multimodal Graph RAG}

\author{Rakesh Raj Madavan*\\
Shiv Nadar University Chennai\\
{\tt\small rakesh21110091@snuchennai.edu.in}
\and
Akshat Kaimal*\\
Shiv Nadar University Chennai\\
{\tt\small akshat21110004@snuchennai.edu.in}
\and
Hashim Faisal*\\
Shiv Nadar University Chennai\\
{\tt\small hashim21110365@snuchennai.edu.in}
\and
Chandrakala S\\
Shiv Nadar University Chennai\\
{\tt\small chandrakalas@snuchennai.edu.in}
}

\begin{document}
\maketitle
\begin{figure*}[ht]
  \includegraphics[width=1\textwidth]{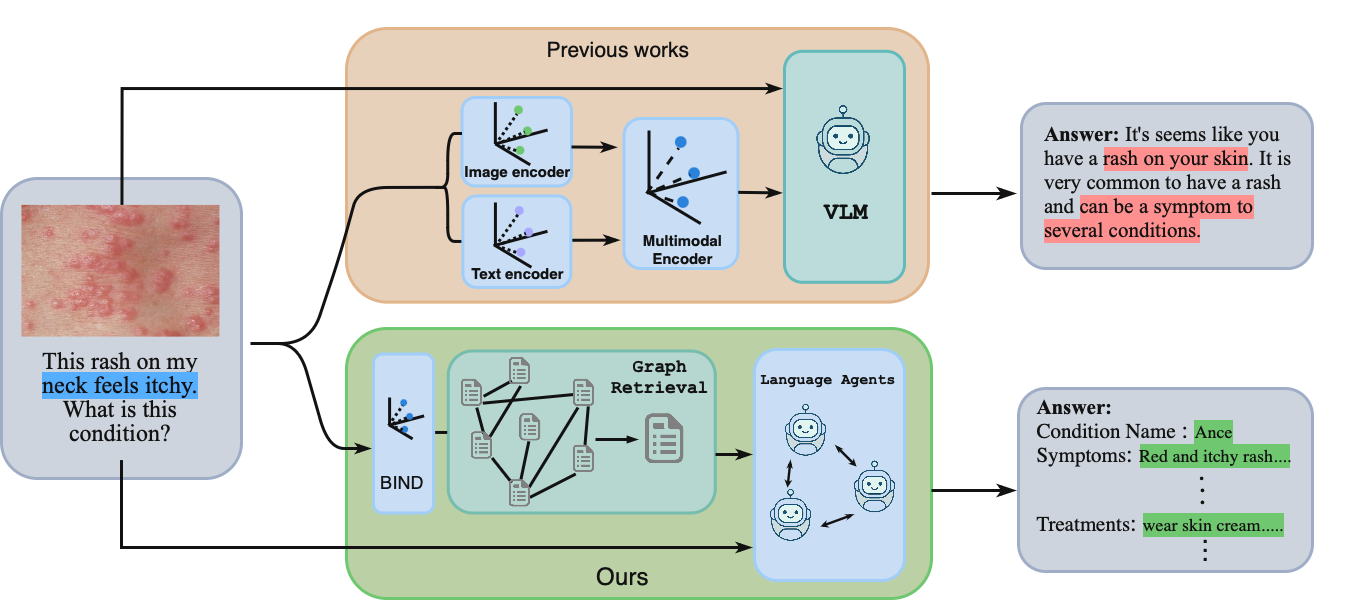}
  \caption{Conveying all the details for medical diagnoses poses significant challenges. To tackle this, we propose Med-GRIM, a novel framework that provides an interactive and detailed response. Existing approaches address this task by directly fine-tuning Vision-Language Models on medical VQA datasets, resulting in sub-optimal responses. In contrast, Med-GRIM integrates a graph-structured datasets and language models, enabling superior performance.
  }
  \label{fig:teaser}
\end{figure*}
\begin{abstract}
An ensemble of trained multimodal encoders and Vision-Language Models (VLMs) has become a standard approach for Visual Question Answering (VQA) tasks. However, such naive models often fail to produce responses with the detailed precision necessary for complex, domain-specific applications such as medical VQA. Our representation model, \textbf{BIND}: \textbf{B}LIVA \textbf{In}tegrated with \textbf{D}ense Encoding, extends prior multimodal work by refining the joint embedding space through dense, query-token-based encodings, inspired by contrastive pretraining techniques. This refined encoder powers \textbf{Med-GRIM}, a model designed for medical VQA tasks that leverages graph-based retrieval and prompt engineering to integrate domain-specific knowledge. Rather than relying on compute-heavy fine-tuning of vision and language models on specific datasets, Med-GRIM applies a low-compute, modular workflow with small language models (SLMs) for efficiency. Med-GRIM employs prompt-based retrieval to dynamically inject relevant knowledge, ensuring both accuracy and robustness in its responses. By assigning distinct roles to each agent within the VQA system, Med-GRIM achieves large language model performance at a fraction of the computational cost. Additionally, to support scalable research in zero-shot multimodal medical applications, we introduce DermaGraph, a novel Graph-RAG dataset comprising diverse dermatological conditions. This dataset facilitates both multimodal and unimodal querying. The code and dataset can be found here \href{https://github.com/Rakesh-123-cryp/Med-GRIM.git}{link}.

\end{abstract}
\section{Introduction}
Multimodal representation learning, inspired by the natural sensory integration of human perception, has gained significant traction in recent years. The increasing availability of large-scale datasets in image, text, and audio modalities, coupled with advancements in generative pretraining, has accelerated the development of versatile encoders capable of understanding and processing multiple modalities effectively \cite{vaswani2017attention,shaikh2023maivar}. One prominent application of multimodal representation learning is Visual Question Answering (VQA), a task that requires a model to comprehend and integrate information from both visual and textual modalities to answer questions about an image \cite{anderson2018bottom,antol2015vqa}. This task goes beyond simple image recognition, challenging models to analyze complex visual cues and extract relevant knowledge to generate accurate answers. 

Among the diverse applications of VQA, medical VQA stands out as a particularly impactful use case. It seeks to aid clinical decision-making by acting as a virtual assistant, capable of interpreting medical images and providing answers to diagnostic questions. However, many existing VQA models struggle to construct an efficient joint embedding space. This inefficiency often arises from the arbitrary nature of multimodal representations learned during pretraining, which can result in poorly aligned embeddings that limit the model's interpretative capabilities. To address this, we introduce BIND, a dedicated mechanism that refines and shifts the learned embeddings into an enhanced space. End-to-end medical vision-language models (VLMs) typically employ instruction fine-tuning on medical VQA datasets. While these models perform adequately on simpler questions, they often fail in harmonizing diverse modality-specific features into a single latent space, which affects the model’s ability to generate contextually accurate answers for more complex or realistic prompts. In contrast, BIND enhances this integration by refining and re-aligning the multimodal embeddings into a cohesive representation space. This approach ensures that the model captures the intricate interdependencies between visual and textual data facilitating a higher level of interpretive accuracy.
\\
 Adapting VQA to knowledge-intensive fields such as medical assessment pose unique challenges, including limited labeled data, domain shifts, and the need for models to generalize across diverse medical datasets. Recent advancements include the development of synthetic data generation techniques\cite{al2020inception, nguyen2019overcoming}, which help address the scarcity of high-quality labeled data by augmenting datasets with artificially generated, medically relevant questions and answers. Additionally, they also lack the level of detail required for medical applications, which we argue extends to the task of \textit{medical querying}. Unlike general purpose VQA, knowledge-intensive fields such as medicine, requires additional context that has to be provided to the model along with domain-specific fine-tuning. Recent efforts\cite{lewis2020retrieval,han2021prompt} have addressed this issue by incorporating Retrieval Augmented Generation(RAG) or Prompt-based approaches that leverage retrieved textual and multimodal examples, respectively, to guide zero-shot response generation, allowing for more open-ended, contextually appropriate answers. Despite rapid advancements, current models typically provide only a single diagnosis, limiting their utility in cases where multiple potential diagnoses should be considered. This approach restricts users from understanding a broader range of possible conditions, especially when the input data is ambiguous or when symptoms may overlap across different medical scenarios. To address this gap, we introduce \textbf{Med-GRIM}, an innovative agentic medical LLM designed to operate within a multimodal GraphRAG\cite{edge2025localglobalgraphrag} framework. Med-GRIM processes multimodal data by embedding an image with the prompt/description provided and dynamically adapts to input ambiguity by retrieving relevant data and engaging in a guided, iterative filtering process. This approach allows the model to respond to user queries with insights on potential conditions, refining its answers by posing clarifying questions and suggesting a range of possible diagnoses.

In response to the growing integration of Explainable AI (XAI) in healthcare, where transparency is increasingly prioritized to clarify how diagnoses and recommendations are made, we propose an approach that embeds explainability by presenting the reasoning and analytical steps underpinning each conclusion. This design offers users clear insights into Med-GRIM's decision-making process, fostering trust and comprehension in AI-driven healthcare solutions.

Our primary contributions are as follows:
\begin{itemize}
    \item We propose a novel multimodal representation learning architecture, \textbf{BIND}, that advances performance on state-of-the-art datasets in multimodal learning.
    \item We introduce \textbf{DermaGraph}, a multimodal, graph-structured dataset designed for RAG tasks in dermatology, which also supports unimodal input scenarios.
    \item We develop a zero-shot learning pipeline specifically tailored for medical query processing, enabling accurate and efficient responses to medical inquiries.
    \item We design a graph filtering mechanism for improved diagnostic accuracy in medical applications.
\end{itemize}

\section{Related Works}

\subsection{Multimodal Representation Learning}

Recent advances in multimodal learning have focused on developing architectures that can effectively bridge the gap between different modalities. Some works \cite{girdhar2023imagebind, zhang2024multimodal} attempt to map between audio, text and image using pre-trained unimodal encoders. When narrowed down to images and text alone, these approaches broadly fall into two categories: (1) Models utilizing learned query embeddings. (2) Models employing joint training strategies.\\

\textbf{Query-Based Multimodal Encoders.}
Several prominent models have adopted query-based approaches for multimodal encoding. An early work in this domain \cite{alayrac2022flamingo} set the foundation, followed by subsequent models that introduced incremental improvements such as BLIP\cite{li2022blip}, Frozen\cite{tsimpoukelli2021multimodal} and VisualGPT\cite{chen2022visualgpt}. MiniGPT-4\cite{zhu2023minigpt} leverages the frozen Q-Former architecture from BLIP2\cite{li2023blip} alongside CLIP\cite{radford2021learning}. Similarly, BLIVA\cite{hu2024bliva} and Veagle\cite{chawlaveagle} extend this approach by incorporating additional image features as direct input to the language model, enhancing the model's visual understanding capabilities. These approaches, while effective, often struggle to fully capture the nuanced relationships between visual and textual modalities due to the inherent limitations of query-based representations.\\

\textbf{Joint Training Approaches.}
An alternative paradigm involves jointly training language models with both textual and visual features. LLaVA\cite{liu2024visual} employed a two-stage protocol: vision-language alignment pre-training followed by visual instruction tuning. Their architecture, notably simple yet effective, combines a pre-trained visual backbone with a large language model (LLM) through a vision-language cross-modal connector. More recent work, such as Phi-3.5-Vision, utilizes CLIP ViT-L/14 for image encoding paired with a Phi-3-4k decoder, demonstrating the potential of end-to-end joint training. While this approach has shown promise, particularly in instruction-following tasks, it faces challenges related to the degradation of the LLM's previously acquired knowledge. Additionally, recent works have explored contrastive learning \cite{zolfaghari2021crossclr}, which applies inter and intra-modality loss functions to align the unimodal encoders with the language model, ensuring semantically consistent representations across modalities.\\
Current models, whether query-based or jointly trained, often learn representations through queries, resulting in sub-optimal embedding spaces. Our work addresses this fundamental challenge by introducing a novel transformation approach that maps processed features into a true joint space. This design choice is supported by our empirical findings, which demonstrate that linear transformation layers fail to match the performance of our proposed architecture.

\subsection{Medical VQA and chatbots}
Most existing Medical VQA typically employs pre-existing architectures that are fine-tuned on medical data. \cite{hu2024interpretable} developed a report generation model focused on detecting regions of interest (ROIs) and linking them using a graph-based structure. The concept of Generalized Medical Artificial Intelligence (GMAI) was introduced in \cite{moor2023foundation}, with further exploration of vision-language models in \cite{zhang2024generalist}. To improve precision in Medical VQA (MVQA) models, RAG has also been widely utilized. For example, \cite{wu2024medical} proposed a method that organizes textual medical information within a hierarchical graph structure, while \cite{xia2024mmed,xia2024rule} employ multimodal prompt retrieval to fetch relevant case examples for a given query. RAMM \cite{yuan2023ramm} takes a similar retrieval-augmented approach, but applies it across both image and text data. Beyond VQA, an interactive agent could greatly enhance user experience by enabling detailed conversations. Currently, there are only a few medical chatbots capable of such engagement, with models like Med-Alpaca \cite{han2023medalpaca}, LLaVa-Med \cite{li2024llava}, and Idefic-Med being among the few that can respond to realistic medical images. However, these models also face limitations, often providing insufficient detail or generating incorrect results when encountering ambiguous inputs.

 \begin{figure*}[ht]
  \includegraphics[width=1\textwidth]{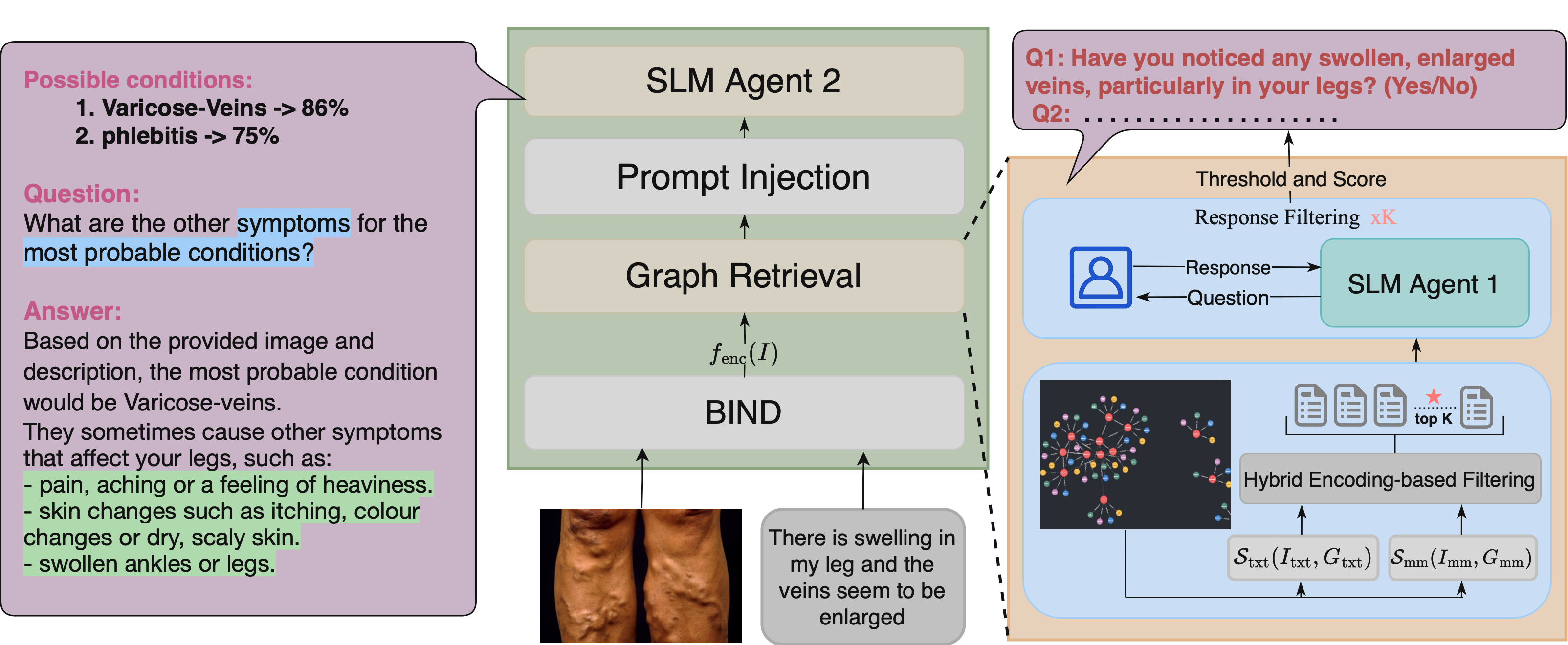}
  \caption{Pipeline of the proposed Med-GRIM system. Med-GRIM integrates multimodal inputs—such as images and descriptions—through a series of specialized modules, including BIND, graph retrieval layers and prompt injection. The model first assesses possible conditions, ranking them by probability, then dynamically retrieves relevant data and refines responses iteratively. This approach allows it to present condition-agnostic insights and tailor responses based on user feedback. Through iterative filtering, Med-GRIM engages users with clarifying questions, adapting its answers based on specific input cues, as shown in the step-by-step reasoning for diagnosing conditions like varicose veins.}
  \label{fig:main_arch}
\end{figure*}

\section{Approach}
\subsection{Architecture Overview}
Med-GRIM operates as a three-stage pipeline integrating our BIND encoder with graph-based retrieval and specialized SLMs. \textbf{Stage 1: Multimodal Encoding} Given input image $I_{img} \in \mathbb{R}^{H \times W \times 3}$ and text prompt $I_{text}$, BIND creates unified representations $I_{mm}$ using our True Transformation Layer (TTL) for enhanced cross-modal relationships. \textbf{Stage 2: Graph-based Retrieval} The encoded query $I_{mm}$ is matched against DermaGraph using two-tier filtering: hybrid encoding-based similarity scoring (Eq. 1-2) followed by SLM-based response filtering with Phi3-3.8B and Mistral-7B agents that generate clarifying questions and assign likelihood scores. \textbf{Stage 3: Response Generation} Filtered conditions are processed by SLM agents to generate comprehensive responses while maintaining conversational state for follow-up questions. The key innovation is the tight integration between BIND's embeddings, graph semantic relationships, and specialized SLM roles, creating synergistic multimodal understanding rather than separate components.

\subsection{BIND}
Previous approaches to learning joint representations rely on queries, analogous to the [CLS] token used in language models such as BERT. While these methods are effective at capturing the semantic meaning of inputs, they often fail to accurately model the vector proximity between embeddings, resulting in a suboptimal embedding space. To address this limitation, we propose the True Transformation Layer (TTL), designed to learn a more robust joint space.

\begin{figure*}[h]
  \centering
  \includegraphics[width=0.85\textwidth]{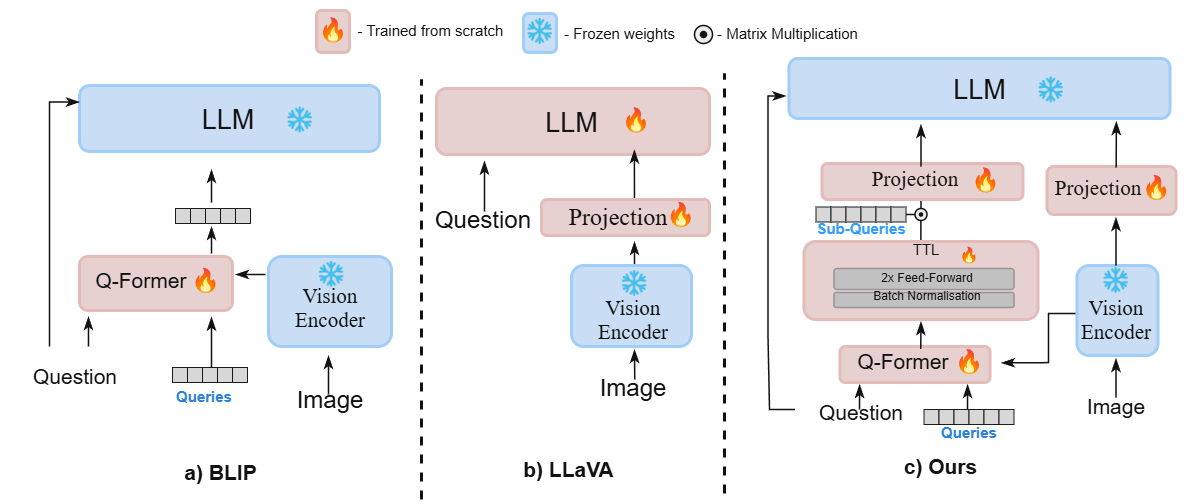}
  \caption{\textbf{Comparison of various VLM architectures:} a)BLIP uses a small set of query embeddings to compress visual information b) LLaVA trains the LLM to understand visual cues c) BIND(Ours) learns representations through sub-query projections and the True Transformation layer(TTL) proposed.}
  \label{fig:MMRL}
\end{figure*}

As illustrated in Fig\ref{fig:MMRL}, we encode the images using a pre-trained feature extractor to ensure intra-modal alignment. For initial inter-modal alignment, we employ the Q-Former model from BLIVA\cite{hu2024bliva}, a BERT-Base decoder that processes both text and image features. Using the TTL layer and the learnable Sub-Query parameter, the model can learn more complex mappings between modalities, potentially creating a more semantically meaningful and geometrically coherent joint embedding space.

After training BIND on a large dataset, we leverage its encoder to encode all multimodal data in our approach.
\subsection{Graph Retrieval}
\label{section:graphdata}
The graph retrieval process is designed to extract information from a dataset structured as a graph. It operates through a dual-stage filtering system to enhance accuracy and relevance: (1) Hybrid Encoding-based filtering, which eliminates dissimilar conditions; (2) Response filtering, where questions are posed to the user based on the filtered conditions to achieve finer filtration. This RAG technique helps prevent hallucinations in the language model, a critical step to avoid presenting incorrect information to the user. Additionally, RAG also removes the need for task-specific fine-tuning of the language model, thus saving substantial time and computational resources.\\

\textbf{Hybrid Encoding-based Filtering. }
We utilize our multimodal encoder \( f_{\text{enc}}(x) \) to measure the connectivity between the user input \( I \) and the data in graph $\mathcal{G}$ (detailed in Sec.\ref{section:Dataset}).
\begin{equation}
\mathcal{S}(x,G) = \left(\frac{f_{\text{enc}}(x) \cdot f_{\text{enc}}(G)}{|f_{\text{enc}}(x)| |f_{\text{enc}}(G)|}\right), where \; G \in \mathcal{G}\\
\end{equation}
\begin{equation}
    \mathcal{O}_\text{t}(I) = \lambda \mathcal{S}(I_\text{text},G_\text{text}) + (1-\lambda) \mathcal{S}(I_\text{mm},G_\text{mm})
    \label{eq:similarity}
\end{equation}
In this equation, $\lambda \in [0,1]$ is a hyperparameter that controls the relative weight of the text-only and multimodal similarities, while $G$ represents all the nodes in the graph $\mathcal{G}$. Alongside the immediate neighbors of the most similar node (represented by the function $neighbors$ explained in Section\ref{section:Dataset}), which introduces additional candidate nodes, we apply adaptive filtering to refine the selection to the top $K$ nodes.
\begin{equation}
    f_{\text{graph}}(x) = \begin{cases} 
    G_{\text{text}} &\text{if }, |x - M| \leq 0.05 \cdot M \\
    \text{None} & \text{otherwise}
    \end{cases}
\end{equation}
\begin{equation}
    where, \;M = \arg\max_{G \in \mathcal{G}} \left(\mathcal{O}_\text{t}(I)\right)
\end{equation}
\begin{equation}
    C = f_{\text{graph}}(\mathcal{O}_\text{t}(I)) \; \cup \; neightbors(f_{\text{graph}}(\mathcal{O}_\text{t}(I)))
\end{equation}

When processing medical data, the input often presents discrepancies, as multiple diseases can exhibit similar symptoms. Text serves as a crucial differentiator in such cases. On the other hand, some diseases may have similar textual descriptions, but their visual representations differ. To address these variations, we introduce a hybrid similarity score, where the text similarity ($\mathcal{S}_\textbf{text}$) and multimodal similarity ($\mathcal{S}_\textbf{mm}$) are combined through a weighted sum, controlled by 
$\lambda$. This approach ensures that both text and image information are effectively considered, enhancing the model’s accuracy in distinguishing between similar conditions. The aforementioned process is formalized in Alg. \ref{alg:graph_filtering1}, where $\mathcal{E}_I$ represents the hybrid encoding of input I.

\textbf{Response Filtering. }
We perform a second stage of filtering where an SLM generates questions based on the symptoms and causes of the conditions $C$ filtered in the previous step. The user’s responses to these questions are recorded and fed back into the SLM, along with their similarity score obtained from Eq.\ref{eq:similarity}, for evaluation through medical reasoning. The SLM then returns a likelihood estimate of the user being affected by each condition. Based on these likelihoods, we filter out conditions with lower probabilities (typically below 50\%).
The algorithmic implementation is delineated under Alg.\ref{alg:graph_filtering2}.\\

\begin{algorithm}
\small
\caption{Stage1-HybridFilter}
\label{alg:graph_filtering1}
\begin{algorithmic}[1]
    \Require Graph $\mathcal{G}$, User input $I$, Weight $\lambda \in [0,1]$
    
    \Function{Stage1-HybridFilter}{$I, \mathcal{G}, \lambda$}
        \State $\mathcal{E}_I = encode(I)$
        \State $scoresList = []$
        \For{each node $G$ in $\mathcal{G}$}
            \State $score \leftarrow \lambda \cdot cos(\mathcal{E}_I^, G) + (1-\lambda) \cdot cos(\mathcal{E}_I, G)$
            \State $scoresList.\text{append}(score)$
        \EndFor
        \State $\text{MaxScore} \leftarrow max(scoresList)$
        \For{each score $score$ in $scoreList$}
            \If{$score \geq 0.95 \cdot \text{MaxScore}$}
                    \State Add $G \cup neighbors(G)$ to candidates
            \EndIf
        \EndFor \\ \;\;\;\;\;\Return candidates 
    \EndFunction
\end{algorithmic}
\end{algorithm}

\begin{algorithm}
\small
\caption{Stage2-ResponseFilter}
\label{alg:graph_filtering2}
\begin{algorithmic}[1]
    \Require Graph $\mathcal{G}$, User input $I$, Weight $\lambda \in [0,1]$
    
    \Function{Stage2-ResponseFilter}{candidates}
        \State questions $\leftarrow$ Generate questions about symptoms
        \State responses $\leftarrow$ Get user responses
        \State filtered $\leftarrow \emptyset$
        
        \For{each condition in candidates}
            \State $prob \leftarrow $EvalLikelihood(condition, responses)
            \If{ $prob >$ 0.5}
                    \State Add condition to filtered
            \EndIf
        \EndFor \\ \;\;\;\;\;\Return filtered 
    \EndFunction
\end{algorithmic}
\end{algorithm}

\begin{table*}[h!]
\centering
\begin{tabular}{lccccccccccc}
\hline
Model & ST VQA & Text VQA & Doc VQA & Info VQA & \textbf{Avg} \\
\hline
OpenFlamingo\cite{alayrac2022flamingo} & 19.32 & 29.08 & 5.05 & 14.99 & 13.67 \\
BLIP2-FLanT5XXL\cite{li2023blip} & 21.38 & 30.62 & 4.00 & 10.17 & 15.00 \\
MiniGPT4\cite{zhu2023minigpt} & 14.02 & 11.52 & 2.97 & 13.32 & 9.58 \\
LLaVA\cite{liu2024visual} & 22.93 & 28.30 & 4.40 & 13.78 & 12.84 \\
InstructBLIP (FlanT5XXL)\cite{li2022blip} & 26.22 & 36.86 & 4.94 & 10.14 & 18.90 \\
BLIVA (FlanT5XXL)\cite{hu2024bliva} & 28.24 & 39.36 & 5.22 & 10.82 & 20.43 \\
\hline
\textbf{BIND (FlanT5XXL)} & \textbf{30.76} & \textbf{42.02} & \textbf{5.21} & \textbf{15.49} & \textbf{23.375} \\
\hline
\end{tabular}
\caption{Zero-Shot evaluation of our method BIND compared to other Vision Language Models on VQA accuracy benchmarks}
\label{tab:MMRL}
\end{table*}

\begin{figure}[h]
  \includegraphics[width=0.45\textwidth]{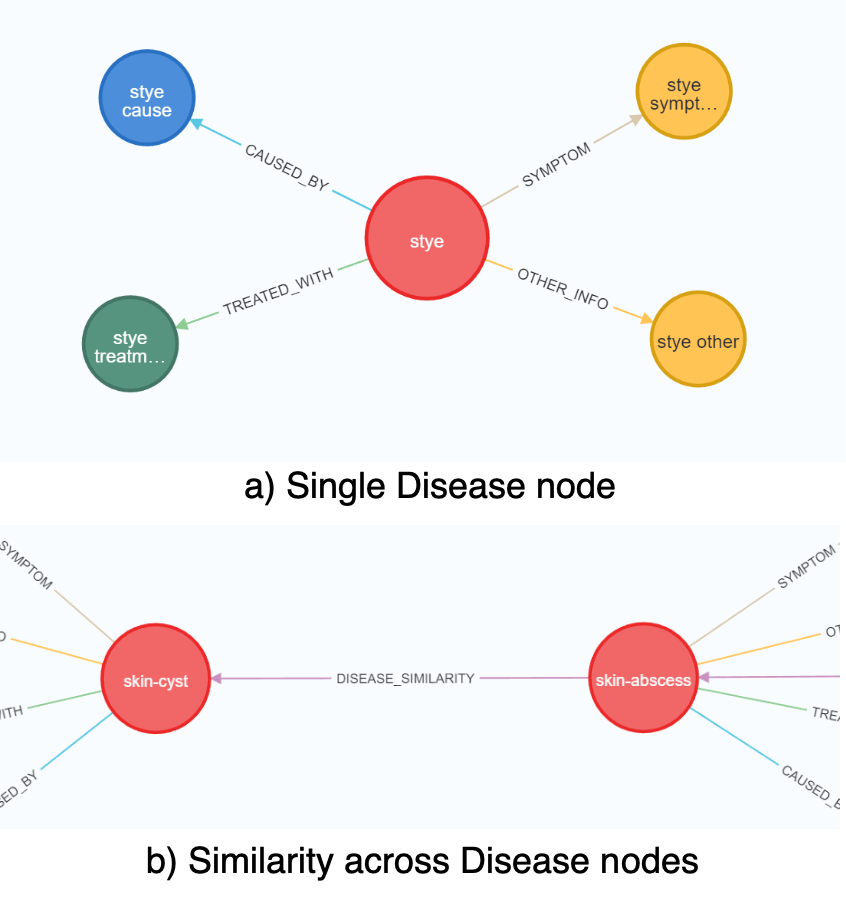}
  \caption{These graph structures depict our dataset DermaGraph at different levels of detail. a) describes a single disease node and b) shows the similarity edge between two disease nodes.}
  \label{fig:dataset}
\end{figure}
\vspace{-10pt}

\subsection{Prompt injection}
The final step, before addressing the user's query, involves integrating the retrieved data into a prompt template to generate a comprehensive response. Prompt engineering has shown promising results in enhancing the quality of language model outputs. It allows for dynamic adaptability, as the prompt template can be modified to suit various query types and domains. To avoid introducing textual artifacts, we design the prompt to direct the language model explicitly toward addressing the user’s concerns, using the text provided through the RAG process.

\section{Experiments and Results}
\subsection{Datasets}
To demonstrate the robustness of our multimodal representation model, BIND, we adopt the training and evaluation methodology outlined in BLIVA \cite{hu2024bliva}. For training, we use a range of datasets, including MSCOCO \cite{mscoco} for image captioning, OCR-VQA \cite{ocr-vqa}, OKVQA \cite{okvqa}, VQAv2 \cite{vqav2}, A-OKVQA \cite{a-okvqa}, and LLaVa-Instruct 150K \cite{llava-instruct}. Evaluation is conducted on a targeted subset, comprising TextVQA \cite{textvqa}, ST-VQA \cite{st-vqa}, Doc-VQA \cite{doc-vqa}, and InfoVQA \cite{infovqa}. Additionally, to assess BIND's performance in the biomedical domain, we incorporate biomedical VQA datasets such as VQA-RAD \cite{vqa-rad} and Path-VQA \cite{path-vqa} for training and evaluation, as detailed in Sec.~\ref{section:exp2}.

\subsection{DermaGraph}
\label{section:Dataset}
To implement RAG for dermatology, we developed a new dataset comprising various dermatological conditions. A total of 50 common dermatological conditions suitable for safe self-diagnosis were sourced from the National Health Service (NHS) website, providing essential details for each condition, including definitions, symptoms, clinical and home treatments, preventive measures, and representative images. We manually curated this data, eliminating redundancies and organizing it into a structured CSV format. To construct the graph database, we utilized Neo4j’s Python API to process the CSV file. As shown in Fig\ref{fig:dataset}(a), each red node represents a specific dermatological condition, along with its multimodal embedding obtained from the BIND encoder model, and connects to three child nodes that store related information fields, such as symptoms, treatments, and preventive strategies. For each condition, we incorporated 10-15 representative images, averaging their embeddings from the vision encoder to produce the multimodal embedding used in the graph. In addition to the standard intra-condition edges that link to in-depth information on various categories, we introduced inter-condition edges based on textual similarity, enhancing semantic relationships across nodes. Each graph node includes a function, $neighbors$, which retrieves all adjacent nodes, facilitating rapid information retrieval.
The constructed graph, illustrated in Fig\ref{fig:dataset}(b), highlights these semantic relationships. Notably, since node connections are based solely on language semantics, the dataset is compatible with unimodal tasks (LLMs) without vision capabilities, extending its utility across various LLM applications in dermatology. To create an evaluation dataset, we generated 30 open-ended Q\&A pairs based on the collected conditions.

\begin{figure*}[ht]
  \includegraphics[width=1\textwidth]{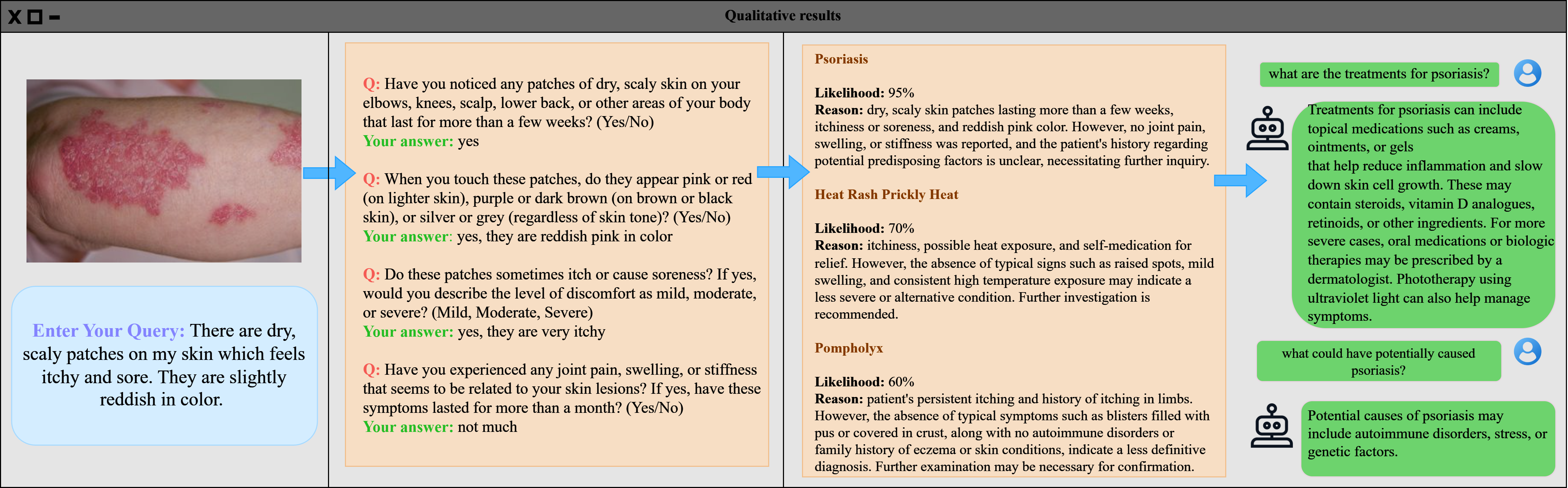}
  \caption{Qualitative analysis of Med-GRIM’s diagnostic flow: From symptom input and follow-up questioning to condition likelihood assessment and interactive Q\&A, Med-GRIM provides a comprehensive, transparent approach to medical diagnoses and user support.}
  \label{fig:qualitative}
\end{figure*}

\subsection{Implementation Details}
We train BIND across all datasets using the AdamW optimizer to minimize cross-entropy loss. The model is pre-trained for two epochs and fine-tuned for 200K iterations on an NVIDIA A40 GPU (48 GB VRAM). We implement a cosine annealing learning rate scheduler with an initial warmup from $10^{-6}$ to $10^{-4}$. The pre-training phase spans two days, while fine-tuning takes four days.
For the SLM agents, we employ Phi3-3.8B as the Q\&A agent and mistral-7B for medical reasoning and user interaction. A $\lambda$ value of 0.4 is used to weight the multimodal embeddings. Inference on a Ryzen 5 4600HS CPU took 17.5 seconds and 6.7 Gb RAM using the mistral-7b LLM, and 13.3 seconds and 5.5 Gb RAM using the Phi-3 LLM along with BIND.\\

\vspace{-15pt}
\subsection{Joint Representation Learning}

\textbf{General VQA.}
We compared our BIND model with other state-of-the-art joint representation learning models, including BLIP, Flamingo, BLIVA, LLaVa, and MiniGPT-4. As shown in Table \ref{tab:MMRL}, our approach demonstrates superior performance on zero-shot VQA benchmark datasets. BIND consistently outperformed other models, underscoring its refined embedding space and alignment across modalities. This performance advantage can be attributed to BIND’s enhanced ability to differentiate embeddings and reduce ambiguity.

\textbf{Biomedical Visual Question Answering.}
\label{section:exp2}
For a domain-specific evaluation, we tested BIND on open-source biomedical VQA datasets, specifically VQA-RAD and PathVQA. As displayed in Table.\ref{tab:biomed} BIND’s performance was compared to prominent Medical Vision-Language Models such as LLaVa-Med, PubMedCLIP, MUMC, BioMedCLIP, and K-Path. Our model exhibited competitive results on both closed-form and open-form query variants, highlighting its adaptability in accurately extracting and correlating image-text features for medical queries. The closed-form tasks required precise image-text understanding to select a predefined answer, while open-form tasks tested the model’s ability to generate responses from scratch. BIND achieved high scores across both formats, reinforcing its suitability for extracting image-text features in Med-GRIM.

\begin{table}
\centering
\small
\begin{tabular}{lcccccccccc}
\toprule
Method & \multicolumn{2}{c}{VQA-RAD(Acc.)} & \multicolumn{2}{c}{PathVQA(Acc.)} \\
           & Open & Closed      & Open & Closed \\
\midrule
\textbf{BIND} & \textbf{72.31} & \textbf{87.5} & \textbf{43.05} & \textbf{95.78} \\
\midrule
MUMC\cite{li2023masked} & 71.5 & 84.2 & 39.0 & 90.1\\
K Path\cite{naseem2023k} & - & - & 42.12 & 94.6 \\
LLaVA\cite{liu2024visual} & 50.00 & 65.07 & 7.74 & 63.20 \\
LLaVA-Med\cite{li2024llava} & 61.52 & 84.19 & 37.95 & 91.21 \\
PubMedCLIP\cite{eslami2023pubmedclip} & - & 80.00 & - & - &\\
BiomedCLIP\cite{zhang2023biomedclip} & - & 79.80 & - & - &\\
\bottomrule
\end{tabular}
\caption{Evaluation of our BIND model against SoTA medical vision-language models fine-tuned on biomedical VQA datasets. BIND outperforms the above methods in both open and closed classification accuracy.}
\label{tab:biomed}
\vspace{-10pt}
\end{table}

\begin{table}[h]
\centering
\small
\begin{tabular}{l@{\hspace{0.5em}}c@{\hspace{0.5em}}c@{\hspace{0.5em}}c}
\toprule
Models & \multicolumn{2}{c}{\textbf{DermaGraph}} \\
& Acc. & Semantic-BERT  \\
\midrule
LLaVA-Med\cite{li2024llava} & 76.7 & 0.63 \\
MUMC\cite{li2023masked} & 80.00 & 0.62 \\
Med-Flamingo\cite{moor2023med} & 46.67 & 0.39 \\
RULE  & 78.5 & 0.75 \\
\midrule
Med-Grim(RAG) & 78.7 & 0.81 \\
\textbf{Med-Grim(Graph-RAG)} & \textbf{83.33} & \textbf{0.81} \\
\bottomrule
\end{tabular}
\caption{Accuracy scores of SoTa multimodal medical language models implemented on DermaGraph. Our model using Graph-RAG outperforms the other methods, including our model using vanilla RAG.}
\label{tab:model_accuracy}
\vspace{-10pt}
\end{table}

\subsection{Prompt-embedded Medical Vision-Language}

We benchmarked Med-GRIM against several multimodal medical LLMs on the DermaGraph dataset, including LLaVA-Med, MUMC, Med-Flamingo and Vanilla RAG methods like RULE. Results in Table \ref{tab:model_accuracy} indicate that Med-GRIM outperformed these models in accuracy across all test cases, highlighting the strength of our prompt-embedded methodology. Additionally, high semantic-BERT scores confirm that Med-GRIM not only delivers accurate responses but also maintains strong coherence. This performance boost is largely due to our two-stage filtering process and the systematic knowledge integration through a graph-based framework.

Performance gains are particularly notable in cases that require detailed reasoning about dermatological conditions, where the model must consider multiple symptoms and visual indicators simultaneously. Our prompt engineering approach, which dynamically injects retrieved knowledge into structured templates, helps maintain consistency while allowing the model to leverage both general medical knowledge and specific dermatological expertise stored in the graph database.
A qualitative example from these experiments is shown in Fig\ref{fig:qualitative}, providing insight into its operation.

\section{Ablation Studies}
In this section, we conduct ablation studies to assess the impact of various parameters and components in our model architecture and approach, providing quantitative results to validate each.

\textbf{Effect of Similarity Weighting Parameter $\lambda$.} The similarity weighting parameter $\lambda$ controls the relative importance of textual versus multimodal input in computing similarity scores. A higher $\lambda$ gives more weight to the textual content, with $\lambda=1$ approximating a conventional language model, and $\lambda=0$ corresponding to an image-to-text translation model. We aim for a balanced integration of text and image features, so extreme values of $\lambda$ were not included in our study. Results for varying $\lambda$ values are summarized in Table \ref{tab:ablation}, demonstrating the influence of this balance on model accuracy.  A $\lambda$ value of 0.4 was found to bear the most accurate results.\\
\begin{table}[h]
\vspace{-10pt}
\centering 
\begin{minipage}{0.45\linewidth} 
  \centering 
  \small
  \begin{tabular}{l@{\hspace{2pt}}c@{\hspace{2pt}}c} 
    \toprule
    $\lambda$ & Ours(Phi-3) & Ours(Mistral) \\
    \midrule
    0.9 & 60.00 & 60.00 \\
    0.7 & 63.3 & 66.7 \\
    0.5 & 70.00 & 76.67 \\
    \textbf{0.4} & \textbf{73.33} & \textbf{76.67} \\
    0.2 & 53.3 & 56.7 \\
    \bottomrule
  \end{tabular}
  \caption{Ablation studies accessing the influence of parameter values across different language models.}
  \label{tab:ablation}
\end{minipage}%
\hfill 
\begin{minipage}{0.45\linewidth} 
  \centering 
  \small
  \begin{tabular}{l@{\hspace{2pt}}c@{\hspace{2pt}}c}
    \toprule
    Model & \multicolumn{2}{c}{Prompt Eng.} \\
    & w/ & w/o \\
    \midrule
    Ours(BLIVA) & 81.91 & 79.1 \\
    Our(BLIP) & 77.5 & 76.67 \\
    \textbf{Ours(BIND)} & \textbf{83.33} & \textbf{80.7} \\
    \bottomrule
  \end{tabular}
  \caption{Accuracy of BIND compared to other medical-VQA models in Med-GRIM with and without implementing prompt engineering}
  \label{tab:prompt_engineering_1}
\end{minipage}
\vspace{-10pt}
\end{table}

\textbf{Influence of Pipeline components.} We tested various language models (LMs) tailored to specific roles within the system, such as response filtering, medical reasoning, and user interaction. Phi-3 performed optimally in the high-parameter range, while Mistral excelled at the 7B parameter level, demonstrating strong zero-shot reasoning capabilities essential for medical querying. In Table \ref{tab:prompt_engineering_1}, we present the comparison of our model BIND with other pre-trained state-of-the-art VLMs in terms of performance in Med-GRIM, with and without prompt engineering. Our model outperforms BLIP and BLIVA, with a noticeable increase in accuracy when using prompt engineering across the board.

\section{Discussion}
\subsection{Future works and Limitations}
Med-GRIM faces challenges due to its dependency on extensively curated data and structured disease datasets maintained by domain experts, potentially limiting scalability. High-dimensional latent representations across modalities may encounter dimensionality issues with current similarity metrics, suggesting the need for more adaptable approaches.

While currently focused on dermatology, Med-GRIM could expand to radiology and pathology through cross-domain graph structures and streamlined Mixture-of-Experts models. To enhance personalization, the system could incorporate user-specific knowledge graphs for tailored medical reasoning.

We've developed a dataset construction pipeline that converts text-image pairs into compatible graph datasets, facilitating expansion beyond the DermaGraph dataset. Our curation process employs an algorithm selecting diverse images based on embeddings, creating a robust dataset focused on common, safely self-diagnosable conditions. Future work will expand both dataset compatibility and disease coverage.

\subsection{Conclusion} 
In this work, we present Med-GRIM, a novel framework for medical VQA that integrates multimodal Graph-RAG with robust quantitative and qualitative filtering strategies, enhancing both precision and efficiency. Existing approaches typically use vision language models (VLMs) for end-to-end response generation or rely on textual retrieval-augmented generation (RAG) systems based on case studies. However, these methods often lack a deep understanding of the given image, provide limited detail, and do not support interactive analysis essential for medical applications. Med-GRIM addresses these limitations by delivering more accurate and detailed responses through user interaction, enabling qualitative filtering of responses. This approach is particularly vital in the medical field, where precision is critical, and it can also be extended to other applications requiring user assistance.
\newpage
{
    \small
    \bibliographystyle{ieeenat_fullname}
    \bibliography{main}
}


\end{document}


\maketitle
\section{Prompt Templates}
Prompt engineering has emerged as a vital technique for improving response generation in large language models (LLMs). It emphasizes structuring and framing instructional prompts to align closely with the patterns and context the model was trained on. Leveraging this approach, we employ a custom-designed prompt template tailored specifically to our application.
Some essential considerations when creating an effective prompt template include:
\begin{itemize}
    \item \textbf{Be as specific as possible: }Providing clear, specific prompts reduces ambiguity, enabling the AI to better understand the query and its context.
    \item \textbf{Specify the level of detail required:} Indicate preferences for the response format, such as using bullet points, specifying the number of questions, or defining paragraph limits.
    \item \textbf{Offer positive guidance:} Focus on explaining what the LLM should do, rather than what it should not do, to maintain clarity and direction.
\end{itemize}

Figure\ref{fig:reason} and Figure \ref{fig:interact} illustrates a sample of the prompt templates used for our agents. The first template demonstrates the structure designed for the agent responsible for medical reasoning, while the second sample outlines the prompt template for the user interaction agent.
\begin{figure}[ht]
  \includegraphics[width=0.5\textwidth]{ICCV2025-Author-Kit-Feb/Images/prompt.png}
  \caption{}
  \label{fig:reason}
\end{figure}
\section{Tip to Use}
Regardless of how well a model performs in response to realistic queries or prompts, it inherently tends to exhibit bias towards the data it was trained on. Publicly available LLMs and VLMs owe their strong performance to the vast and diverse datasets used during training. However, beyond leveraging this pre-existing knowledge, models often require additional, situation-specific details to generate accurate results.

To achieve better performance, the first step is to \textbf{provide a detailed and descriptive prompt about the user's experienced problem}. This enables the model to make informed comparisons and assessments based on the provided context. We also encourage users to \textbf{include any additional symptoms or difficulties not evident in the image}, such as high temperature, headache, or other relevant details.

The quality of the input image is another critical factor influencing the accuracy of the response. A clear, well-focused image of the medical condition can greatly enhance the model's ability to provide an accurate diagnosis. To optimize results, we advise users to \textbf{provide a close-up of the Region of Interest (ROI) while avoiding unnecessary background}. This ensures the model focuses on the medical condition, as the training datasets predominantly consist of images specifically centered on such conditions.
\begin{figure}[ht]
  \includegraphics[width=0.5\textwidth]{ICCV2025-Author-Kit-Feb/Images/prompt_template.png}
  \caption{}
  \label{fig:interact}
\end{figure}
\section{Dataset Samples}
We provide further insights into the dataset developed. Figure \ref{fig:content} illustrates the various attributes contained within the main node for each dermatological condition. Figure \ref{fig:single} presents an enlarged view of the structure of an individual node and also highlights the interconnected multi-node structures described in the paper.
\begin{figure}[ht]
  \includegraphics[width=0.5\textwidth]{ICCV2025-Author-Kit-Feb/Images/content_dataset.jpeg}
  \caption{}
  \label{fig:content}
\end{figure}
These connections illustrate the semantic relationships across nodes, linking dermatological conditions with overlapping attributes, such as shared symptoms, similar underlying causes, or common treatment strategies. For instance, conditions with symptoms like redness or inflammation may be grouped together, creating meaningful clusters that aid in understanding correlations between different conditions. Similarly, treatment strategies that apply to multiple conditions, such as the use of topical steroids or antihistamines, establish additional connections between nodes. This interconnected design transforms the dataset into a dynamic knowledge graph, where the relationships between nodes provide rich context for various tasks.

In applications like differential diagnosis, these semantic links allow the model to compare and contrast conditions, narrowing down possibilities based on the provided input. For retrieval-based reasoning, the graph structure enables the efficient retrieval of related information, offering explanations or additional insights for user queries. This design not only improves the accuracy and robustness of the model’s responses but also provides a foundation for more advanced reasoning tasks, such as identifying rare conditions or suggesting alternative treatments based on related cases. By leveraging these semantic connections, the dataset ensures versatility and adaptability across a wide range of medical and diagnostic applications.
\begin{figure}[ht]
  \includegraphics[width=0.5\textwidth]{ICCV2025-Author-Kit-Feb/Images/single.png}
  \caption{}
  \label{fig:single}
\end{figure}
